# Staged Mixture Modeling and Boosting

Christopher Meek, Bo Thiesson and David Heckerman
Microsoft Research
Redmond, WA 98052-6399
{meek,thiesson,heckerma}@microsoft.com

## Abstract

In this paper, we introduce and evaluate a data-driven staged mixture modeling technique for building density, regression, and classification models. Our basic approach is to sequentially add components to a finite mixture model using the structural expectation maximization (SEM) algorithm. We show that our technique is qualitatively similar to boosting. This correspondence is a natural byproduct of the fact that we use the SEM algorithm to sequentially fit the mixture model. Finally, in our experimental evaluation, we demonstrate the effectiveness of our approach on a variety of prediction and density estimation tasks using real-world data.

## 1 Introduction

In this paper, we introduce and evaluate what we call the staged mixture modeling (SMM) approach: a data-driven staged mixture modeling technique for building density, regression, and classification models. Our approach is to add components to a finite mixture model in stages using the structural expectation maximization (SEM) algorithm. More specifically, at the $n^{th}$ stage, we fix the relative mixture weights and parameters of the first $n-1$ components of the mixture model, and add the $n^{th}$ component with a prespecified initial mixture weight. We then learn the new component and mixture weight using a criterion such as the Bayesian Information Criterion (BIC); a penalized maximum likelihood.

We show that our method is qualitatively similar to a variety of boosting methods. Boosting methods are ensemble methods in which one sequentially adds new predictor components to the ensemble (e.g., Freund & Schapire, 1997). The new predictor components are trained on the basis of a reweighted version of the data set in which cases that are not predicted well are given a higher weight. The connection to boosting is a natural byproduct of the fact that we use the SEM algorithm to sequentially fit the mixture model. Effectively, the SEM algorithm reweights the cases by computing a membership probability for the new component. The membership probability reflects the degree to which the data are not well-predicted by the mixture model without the current component—the worse the prediction, the more weight the case is given. The reweighted data is then used to learn the new component. Although our method is qualitatively similar to many approaches for boosting, it differs in many specific details. We highlight some of the differences by contrasting our approach with the popular boosting methods of Friedman, Hastie, & Tibshirani (1998) and Friedman (1999).

Our approach has several benefits over alternative approaches to boosting. First, our method can easily be applied to any learning method that can learn from fractionally weighted data. Second, our method allows one to boost density models as well as regression and classification models. In addition, our method provides a principled means of optimizing both the weights and the structures of the component models.

In our experimental evaluation, we evaluate the performance of our approach on a variety of prediction and density estimation tasks using real-world data. We use the following types of component models: For classification, we use decision trees with a bounded number of leaves; and for density estimation, we use Bayesian networks whose local distributions are regression trees with a bounded number of leaves. We also evaluate various alternative versions of our algorithm to highlight which aspects are crucial to successful implementation.

## 2 Algorithm

Throughout the paper, we use the following syntactic conventions. We denote a variable by an upper case token (e.g. $A$, $B_i$, $Y$) and a state or value of that variable by the same token in lower case (e.g. $a$, $b_i$, $y$). We denote sets with bold-face capitalized tokens (e.g. $\mathbf{A}$, $\mathbf{X}$) and corresponding sets of values by bold-face lower case tokens (e.g. $\mathbf{a}$, $\mathbf{x}$).

Our approach is based on mixture models. An $n$ component mixture model is a model of the form

$$p^n(\mathbf{y}|\mathbf{x}, \theta) = \sum_{i=1}^{n} p(C = i|\theta_0)\, p_i(\mathbf{y}|C = i, \mathbf{x}, \theta_i)$$

where $\theta$ are the parameters, $p(C = i|\theta_0)$ is the mixture weight of the $i^{th}$ component, and $p_i(\cdot|\cdots)$ is the $i^{th}$ component. For compactness, we will often write $p^n(\cdot)$ for an $n$-component mixture model, $p_i(\cdot)$ for a component model, and $\pi_i$ for the $i^{th}$ component's mixture weight. Special cases of interest are (1) density estimation, in which $\mathbf{X}$ is empty, (2) regression, in which $\mathbf{Y}$ is a single continuous-valued variable, and (3) classification, in which $\mathbf{Y}$ is a single discrete-valued variable. All three of these cases are popular uses of mixture modeling; our methods apply to each of these cases. To simplify the presentation, we assume that the data used to train our model is complete data for $\mathbf{X}$ and $\mathbf{Y}$ (i.e. there is no missing data).

Our approach is a staged approach to constructing a mixture model. At each stage, we add a prespecified initial component to our mixture model with a pre-specified initial mixture weight, while fixing the previous component structures, parameters, and relative mixture weights. We then use a structural expectation maximization (SEM) algorithm to modify the initial component and initial mixture weight in the staged mixture model.

A SEM algorithm is an EM type algorithm in which one computes expected sufficient statistics for potential component models and interleaves structure and parameter search. SEM approaches have been applied to learning of mixtures of Bayesian networks by Thiesson, Meek, Chickering, & Heckerman (1999), to mixtures of trees by Meilă and Jordan (2000), and to the learning of Bayesian networks with missing data by Friedman (1997) who also coined the name.

The concept of a (fractionally) weighted data set for a set of variables is central to the description of our approach. A data set $\mathbf{d} = \{z_1, \ldots, z_N\}$ for a set of variables $\mathbf{Z} = \mathbf{X} \cup \mathbf{Y}$ is a set of cases $z_i$ ($i = 1, \ldots, N$) where $z_i$ is a value for $\mathbf{Z}$. A *weighted case* $wc_i = \{\mathbf{z}_i, w_i\}$ for a set of variables $\mathbf{Z}$ has a value $\mathbf{z}_i$ for the variables $\mathbf{Z}$ and a real-valued weight $w_i$. A *weighted data set for* $\mathbf{Z}$ (denoted $\mathbf{wd} = \{wc_1, \ldots, wc_N\}$) is a set of weighted cases for $\mathbf{Z}$.

In a traditional approach to learning an $n$-component finite mixture model, the E-step of the EM (or SEM) algorithm results in $n$ weighted data sets. If the training data set is $\mathbf{d} = \{z_1, \ldots, z_N\}$, then the weighted data set $\mathbf{wd}_i$ associated with the $i^{th}$ component has weighted cases $wc_j = \{\mathbf{z}_j, p(C = i|\mathbf{z}_j, \theta)\}$ ($j = 1, \ldots, N$) where $\theta$ are the current parameters of the staged-mixture model and $p(C = i|\mathbf{z}_j, \theta)$ is the membership probability for case $j$ in component $i$. We call the quantity $\sum_i w_i$ the *fractional count* for component $i$.

We now describe our algorithm. Its key component is the procedure *Add-Component* that adds a new component to the current mixture model. The procedure takes three arguments: an initial mixture weight $\pi_n$ for the (new) $n^{th}$ component, an initial guess for the $n^{th}$ component $p_n(\cdot)$, and the previous $n-1$ component mixture model $p^{n-1}(\cdot)$. The procedure makes use of two essential routines: (1) a *fractional-data learning method*—a method that can be applied to weighted data set for $\mathbf{X}, \mathbf{Y}$—that produces a probabilistic model for $p(\mathbf{y}|\mathbf{x})$ and (2) a *model score method* that evaluates the fit of a component model to a weighted data set for $\mathbf{X}, \mathbf{Y}$. Note that many fractional-data learning methods employ such a model score (e.g. maximum likelihood, BIC and a Bayesian score).

**Add-component**$(\pi_n, p_n(\cdot), p^{n-1}(\cdot))$

0 Let $p^n(\cdot) = \pi_n p_n(\cdot) + (1 - \pi_n)p^{n-1}(\cdot)$

1 Do $s_1$ steps of structure search

  - use $p^n$ to compute the weighted data set for the $n^{th}$ component.

  - Use weighted data and fractional-data learning method to learn new component $p'_n$

  - if the model score for the new component $p'_n$ on the weighted data does not improve over the model score for the old component $p_n$ on the complete data, then go to step 2.

  - Let $p^n(\cdot) = \pi_n p'_n(\cdot) + (1 - \pi_n)p^{n-1}(\cdot)$

2 Do $s_2$ steps of optimizing mixture weights

  - use $p^n$ to compute the fractional count for the $n^{th}$ component.

  - Perform maximization step for mixture weight to obtain $\pi'_n$

  - let $p^n(\cdot) = \pi'_n p_n(\cdot) + (1 - \pi'_n)p^{n-1}(\cdot)$

3 Repeat step 1 and step 2 $s_3$ times.

4 return $p^n(\cdot)$

In constructing a SMM, we iteratively apply the Add-Component procedure to previously constructed mixture models. We typically construct the first component by applying the fractional-data learning method used in step 1 of the Add-Component procedure to the original (equally weighted) data.

We have found that a good initial model is a marginal model—one in which all variables are assumed to be mutually independent. For regression and classification, a marginal model is simply a univariate marginal distribution of the target variable.

The precise *schedule* of our SEM algorithm is defined by the tuple $(s_1, s_2, s_3)$. Thiesson et al. (1999) have demonstrated that the performance of the learned model is not very sensitive to the precise schedule for an SEM algorithm when applied to mixture modeling, whereas the schedule does affect the runtime of the procedure. We provide additional experiments on alternative schedules in Section 4, and demonstrate that extreme schedules (e.g.) $s_1 = s_2 = s_3 = 1$ can perform poorly.

It is interesting to consider the convergence properties of the Add-Component procedure. Because the EM algorithm is guaranteed to improve the likelihood at each step, if we do no structure search, the Add-Component procedure will improve the log-likelihood on the training data. Similarly, Friedman (1997) showed that the SEM algorithm is guaranteed to improve the overall Bayesian Information Criterion (BIC) if one uses BIC to evaluate the fit of a model to the fractional data during model search. Thus, if we use BIC as a model score in step 1, we can guarantee that the result of the Add-Component procedure will be a local maximum in terms of BIC, if we run to convergence. A similar result holds when using the maximum likelihood (ML) criterion for evaluating the fit of a model.

In addition, if we require that at each stage (i.e., at each application of Add-Component) we only accept the addition of a component to our mixture model if the BIC (ML) score improves, we can guarantee that the SMM approach will identify a parameterized mixture model that is a local maximum in terms of BIC (ML), if we run to convergence. However, it is unclear whether this method is the best method for choosing the number of components in a SMM. First, for mixture models of this type, it is unclear whether BIC is an appropriate score (Geiger, Heckerman, King, & Meek, 2000). Second, we have found that using BIC to select the number of components of a non-SMM mixture model does not yield as good a predictive model as when the number is chosen with holdout data. In this paper, we do not optimize the number of components. Instead, we show that, for the range of numbers of components we consider, our approach roughly monotonically improves performance on a test set, as each component is added.

It is natural to consider variants of the staged mixture modeling approach described above. A natural alternative is to do some type of *backfitting* in which one does not fix the previous components and/or relative mixture weights. In our experiments, we consider two types of backfitting. One, we consider *mixture-weight backfitting* in which we relax the restriction of fixed relative mixture weights. That is, after we have learned and fixed the structure and parameters of each component, use the EM algorithm to estimate the maximum likelihood estimates for all of the mixture weights. Two, we consider *structure backfitting* in which we use the SEM algorithm in conjunction with fractionally weighted data to relearn the structures, parameters, and mixture weights of all components. It is important to note that these alternative approaches typically require more computation than does our SMM approach. The additional computation required is especially large in the case of structure backfitting.

## 3 Relationship to Boosting

In this section, we compare and contrast our approach to constructing mixture models with boosting. We show that our approach to constructing mixture models is qualitatively similar to boosting and distinguish our method from those of Friedman et al. (1998) and Friedman (1998).

When adding the $n^{th}$ component to a mixture model, the weight of the $i^{th}$ case $(\mathbf{x_i}, \mathbf{y_i})$ when initially training the $n^{th}$ component is its membership probability for the case. Recall that we are given an initial mixture weight $\pi_n$ and an initial component $p_n(\cdot)$ as well as our previously constructed $n-1$ component mixture model $p^{n-1}(\cdot)$. The mixture weight for case $i$ is

$$w_i = \frac{\pi_n p_n(\mathbf{y_i}|\mathbf{x_i})}{\pi_n p_n(\mathbf{y_i}|\mathbf{x_i}) + (1-\pi_n)p^{n-1}(\mathbf{y_i}|\mathbf{x_i})}.$$

When using the maximum likelihood or BIC approach for training, what is important in understanding the effect of reweighting the data is the relative size of the mixture weights across cases. We consider two cases $i$ and $j$, and simplify the analysis by assuming that each initial $p_n(\mathbf{y_i}|\mathbf{x_i})$ is a uniform distribution. (The analysis of the relative mixture weights when non-uniform initial components $p_n(\cdot)$ are used is more complicated but qualitatively similar.) Under this assumption, the ratio of the mixture weights for case $i$ over case $j$ is

given by

$$\frac{w_i}{w_j} = \frac{\pi_n p_n(\cdot) + (1-\pi_n) p^{n-1}(\mathbf{y_j}|\mathbf{x_j})}{\pi_n p_n(\cdot) + (1-\pi_n) p^{n-1}(\mathbf{y_i}|\mathbf{x_i})}.$$

Consequently, if case $j$ is better predicted than is case $i$ by the $n-1$ component model, then the mixture weight ratio is larger than one. Furthermore, the better case $j$ is predicted, the larger the ratio. Thus, cases that are poorly predicted by the $n-1$ component model are given relatively larger weights. Also, we can amplify weight differences between cases by increasing the initial mixture weight $\pi_n$.

We have demonstrated that our approach is qualitatively similar to other approaches to boosting in that we more heavily weight cases that are poorly predicted by the previous ensemble of components. Now we compare our approach to other boosting approaches to highlight significant differences.

In many approaches to boosting, including those of Friedman et al. (1998) and Friedman (1999), the components of the ensemble are combined with both positive and negative weights. In our approach, because we are constructing a mixture model, only positive weights are used. Another significant difference between our approach and other boosting approaches is the form of the model. For instance, in the case of classification, our probability estimate of a target class is a linear combination of the probability estimates for the components. In the gradient boosting approach of Friedman (1999) and the LogitBoost approach of Friedman et al. (1998), it is the log odds ratio that is a linear combination of the outputs of the components of the ensembles. Another distinguishing feature of our approach is that, due to the use of EM, we can, at a given stage, iteratively reweigh the data to optimize both the component structure parameterization and mixture weight (i.e., we can set each $s_i$ to be greater than 1). Other approaches such as Friedman's, typically only perform a single line search to obtain the combination weight and do not reweigh the data in the process of constructing the new component in the ensemble. In the next section, we demonstrate that departing from the boosting-like schedule $s_1 = s_2 = s_3 = 1$ typically improves the performance of our approach.

## 4 Experiments

In this section, we describe our experimental results of applying the staged mixture modeling approach to density estimation and classification problems.

| Group | Name | #Train | #Test | #Vars |
|---|---|---|---|---|
| Digits | Digit 0 | 1100 | 434 | 64 |
| | Digit 1 | 1100 | 345 | 64 |
| | Digit 2 | 1100 | 296 | 64 |
| | Digit 3 | 1100 | 260 | 64 |
| | Digit 4 | 1100 | 234 | 64 |
| | Digit 5 | 1100 | 193 | 64 |
| | Digit 6 | 1100 | 281 | 64 |
| | Digit 7 | 1100 | 241 | 64 |
| | Digit 8 | 1100 | 216 | 64 |
| | Digit 9 | 1100 | 211 | 64 |
| Speech | M54 | 1560 | 14 | 33 |
| | M56 | 2336 | 52 | 33 |
| | M64 | 1659 | 9 | 33 |
| | M78 | 6294 | 73 | 33 |
| | N86 | 8688 | 98 | 33 |
| | N99 | 10127 | 227 | 33 |
| | N146 | 4791 | 69 | 33 |
| | N158 | 1796 | 21 | 33 |
| | Z134 | 21888 | 4378 | 33 |

| Group | Name | #Train | #Test | #Vars | #Classes |
|---|---|---|---|---|---|
| UCI | Vowel | 528 | 462 | 10 | 11 |
| | Satimage | 4435 | 2000 | 36 | 6 |
| | Letter | 16000 | 4000 | 16 | 26 |

Table 1: Statistics of the data sets used in our experiments.

### 4.1 Data Sets

In our experiments, we use three groups of data sets: Digits, Speech, and UCI. The first two groups are used to evaluate the performance of our staged mixture modeling approach on the task of density estimation, and the third is used to evaluate the task of probabilistic classification. Characteristics for the data sets are summarized in Table 1.

The first group, Digits, are digital gray-scale images of handwritten digits made available by the US Postal Service Office for Advanced Technology (Hinton, Dayan, & Revow, 1997). The second group of data sets, Speech, contains data sets for individual sub-phonetic events observed for 10ms time frames of continuous speech (Huang et al., 1995). The third group, UCI, contains benchmark data sets from the UCI repository. We chose three data sets for this group—Vowel, Satimage, and Letter—based on ability to use the same training and test data as used in Friedman, Hastie, & Tibshirani (1998).

### 4.2 Models

In our density estimation experiments, the component models of our staged mixture models are Bayesian networks in which each local distribution is a regression tree. For our classification experiments, our component models are single decision trees.

As in the approaches of Friedman et al. (1998) and Friedman (1999), we restrict the maximum number of leaves and use the maximum likelihood criterion when constructing our regression/decision trees. We use the standard greedy search approach for construction except that, as described in Chickering *et al.* (2001), we consider seven split points for a continuous input/regressor variable, a choice that we have found to be a good one over a wide variety of data sets. In the case of density estimation, we enforce the acyclicity constraint of the Bayesian network at each stage of the construction (see Chickering, Heckerman, & Meek, 1997). We choose the maximum number of leaves and our initial mixture weight ($\pi_n$) using a 70/30 split of the training data. In our experiments, we use the Bayesian information criterion (BIC) as our model score in step 1 and chose $s_1 = 5$ and $s_2 = 5$. With respect to the schedule parameter $s_3$, we run until convergence or a maximum of 20 iterations (5 iterations for tuning experiments), whichever occurs first. We say that convergence is reached if the difference in the log-likelihood of the model after step 1 and step 3 divided by the difference in the log-likelihood of the model after step 3 and the initial model falls below $10^{-5}$.

We compare our staged mixture models for density estimation to a baseline model that is a single Bayesian network in which each local distribution is a regression tree. Similarly, for classification, we compare with a single decision tree. The baseline models are learned as are the components of the SMM, except that we do not restrict the maximum number of leaves and we use a Bayesian score to construct the tree. In our Bayesian score, we use a non-informative prior distribution for the parameters in all leaf distributions and a structure prior proportional to $\kappa^d$ where $d$ is the number of free parameters in the model. In a non-Bayesian fashion, we tune the parameter $\kappa$, and the parameter $\gamma$—the minimum number of observations required for a split—using a 70/30 split of the training data. For more details on the Bayesian score for regression trees, see Meek, Chickering, & Heckerman (2002); and for decision trees see Chickering *et al.* (1997).

We measure the performance of structured mixture models and our baseline models with the following measures. For Digits and Speech—our density estimation data sets—we measure the quality of learned models by using the *log-score* on a test set $\mathbf{t} = (\mathbf{y_1}, \ldots, \mathbf{y_N})$ of $N$ cases:

$$Score(\mathbf{t}|model) = 1/N \sum_{i=1}^{N} \ln p(\mathbf{y_i}|model).$$

For the UCI data sets—our classification regression data sets—we measure the log-score for target given input on a test set $\mathbf{t} = ((y_1, \mathbf{x_1}), \ldots, (y_N, \mathbf{x_N}))$:

$$Score(\mathbf{t}|model) = 1/N \sum_{i=1}^{N} \ln p(y_i|\mathbf{x_i}, model).$$

We also evaluate the accuracy of our method as compared to the boosting method of Friedman *et al.* (1998) for the UCI data sets. We measure the classification accuracy:

$$Acc(\mathbf{t}|model) = 1/N \sum_{i=1}^{N} \chi^{y_i}(\arg\max_{y} p(y|\mathbf{x_i}, model))$$

where $\chi^{y_i}(y)$ is 1 if $y_i = y$ and is 0 otherwise.

### 4.3 Results

The log-scores of SMMs and baseline models are reported in Figure 1. Each graph depicts the results for one of the data sets. We show only three graphs for the Digits data sets; for the digits "1", "2", and "3". These results are representative of all of the results on the Digits data sets.

The graphs demonstrate that the SMM approach yields models with good predictive performance. This is true for both density modeling tasks (Digits and Speech data sets) as well as for classification tasks (UCI data sets). In all but three cases, the SMM models obtain better log-scores than the baseline models. For all but one of the data sets, we see that the SMM model achieves the same or better results than the baseline model. For all data sets, the SMM approach rapidly improves on the initial single component model, although, as one would expect, the rate of improvement decreases as additional components are added. The most extreme examples of this pattern are found in the Digit data sets, where improvement is slow or non-existent after the addition of the second component.

Our results on classification accuracy are presented in Table 2. We present results for SMMs with 16 components and LogitBoost models with 200 components; the choice of 200 components yields the best performance for LogitBoost. Note that accurate SMMs have far fewer components than accurate LogitBoost models. Although the SMM component models are more complicated than those for LogitBoost models, we suspect that the difference is in part due to the fact that we iteratively reweight the data for a component to optimize the mixture weight and component (i.e. $s_i > 1$), whereas LogitBoost does not. Our results for classification accuracy on the UCI data sets are mixed. On Vowel, our method performs slightly better than either versions of LogitBoost; on Satimage and Letter, our method performs slightly worse. We attribute these observations to several factors. First, the model class used by LogitBoost is different. Second, LogitBoost regularizes mixture weights as new components are added, whereas we are using maximum likelihood to add new components.

| Data set | Baseline | SMM 16 | LB(2) 200 | LB(8) 200 |
|---|---|---|---|---|
| Vowel | 0.431 | 0.491 (16) | 0.489 | 0.483 |
| Satimage | 0.851 | 0.883 (128) | 0.898 | 0.912 |
| Letter | 0.863 | 0.906 (512) | 0.855 | 0.967 |

Table 2: Classification accuracy for baseline model, SMM with 16 components, LogitBoost model with 200 two-leaf components, and LogitBoost model with 200 eight-leaf components. The number of leaves used in the decision trees for the components of the SMM is given in parentheses after the accuracy.

Next, let us examine the sensitivity of the results to various algorithm parameters. One parameter in the SMM approach is the initial mixture weight $\pi_n$. The quality of the learned models measured in log-score is only moderately sensitive to the choice of $\pi_n$. Representative of all data sets we have examined, Figure 2 plots the log-score on the test set as a function of $\pi_n$ and the number of components in the mixture for the UCI data set Satimage.

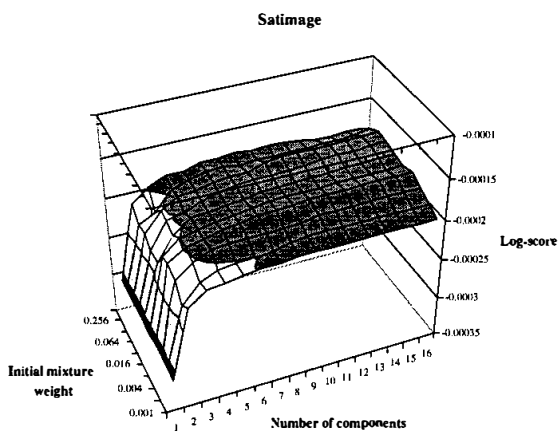

Figure 2: Log-score as a function of initial mixture weight for new component and the number of components in mixture.

In addition, there are (at least) two natural candidates for an initial model: marginal (described above) and uniform. In experiments not reported here, we have found that the accuracy of classification is not sensitive to this choice, whereas the accuracy of density estimation is better when marginal models are used.

Now let us consider the backfitting alternatives described in Section 2: mixture-weight backfitting and structure backfitting. The results on two representative data sets are given in Figure 3. Each plot represents the log-score as a function of number of components for SMM, mixture-weight backfitting, and structure backfitting. For Letter, we see that structure backfitting hurts performance and mixture-weight backfitting hurts performance to a lesser degree. The results for the Speech data set N146 are similar except that the structure backfitting not only hurts performance but additional components significantly degrade performance. The predictive performance of backfitting methods on other data sets is roughly evenly split between these two types of behaviors. In a small number of experiments, we found mixture-weight backfitting adversely affected performance. In summary, our staged mixture modeling approach can both improve predictive performance and reduce computational cost.

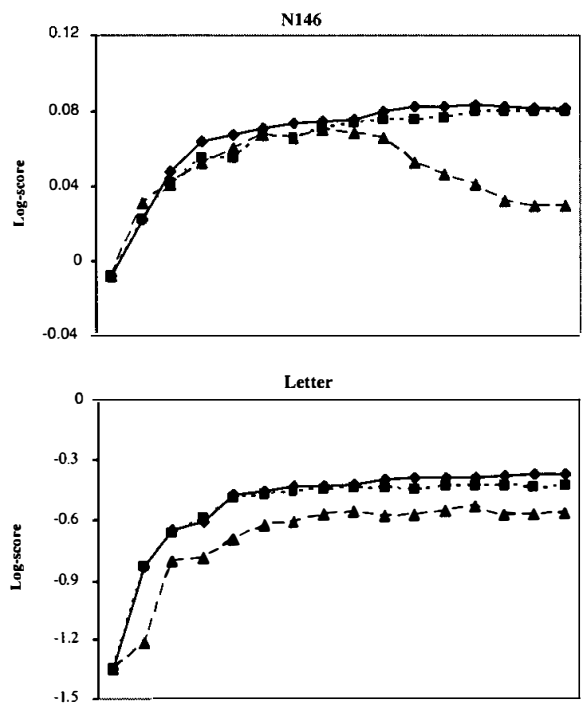

Figure 3: Performance of backfitting for mixture models with 1 to 16 components for N146 (top) and Letter (bottom). Lines labeled with diamonds, squares, and triangles correspond to SMM (no backfitting), mixture-weight backfitting, and structure backfitting, respectively.

Next, let us consider variations in the schedule parameters $s_1$, $s_2$, and $s_3$. Figure 4 shows log score as a function of number of components in the learned mixture model for four different schedules applied to the the Letter and N146 data sets. We use "SMM" to denote the schedule used in most of our experiments $s_1 = 5$, $s_2 = 5$, $s_3 = 20$, "20-1-1" to denote the schedule $s_1 = 20$, $s_2 = 1$, $s_3 = 1$ in which we perform 20 structural searches and one weight update, "1-20-1" to denote the schedule $s_1 = 1$, $s_2 = 20$, $s_3 = 1$ in which we perform 1 structural search and 20 weight updates, and "1-1-1" to denote the schedule $s_1 = 1$, $s_2 = 1$, $s_3 = 1$. The plots in Figure 4 are representative of the performance of these schedules on the other data sets. In both plots, the performance of the schedules "1-1-1" and "1-20-1" are worse than the schedules "SMM" and "20-1-1". This observation suggests that allowing additional steps of structural search while adding new components—a divergence from the boosting-line schedule—is important for improving performance.

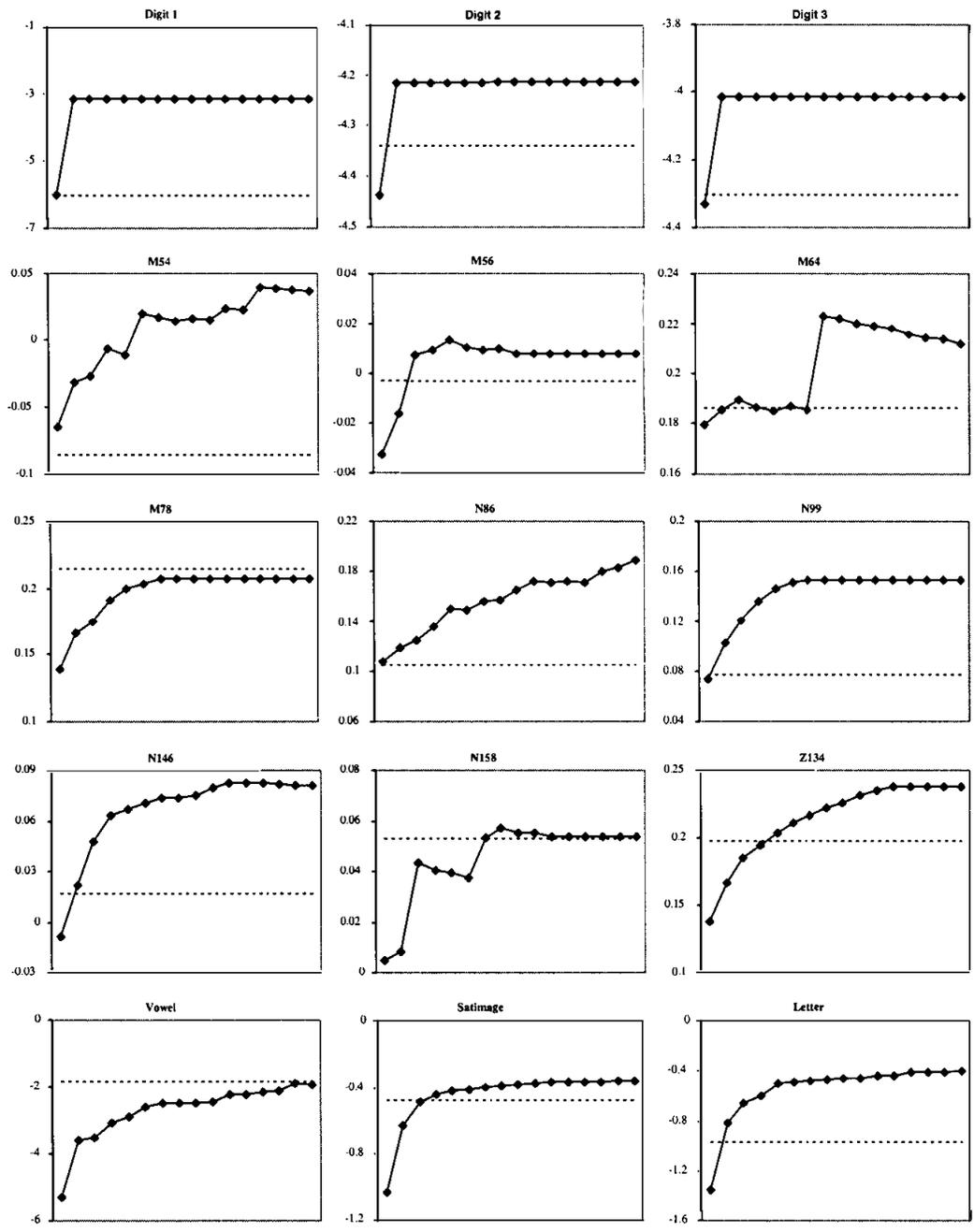

Figure 1: Log-scores on test sets for SMM with 1 to 16 components. Log-scores of the baseline models are shown as horizontal dotted lines.

Finally, let us consider how prediction accuracy is affected by the addition of many model components. Results are again shown in Figure 4. For no schedule does performance systematically degrade as we increase the number of components in the mixture models. This observation suggests that our SMM approach to constructing mixture models is robust to overfitting.

## 5 Discussion

We described our staged mixture modeling approach and provided experimental evidence that it yields high-quality predictive models. We demonstrated that we can improve the quality of both density and classification models using this approach. One of the benefits of the SMM approach is that it can be used with any component model that can be learned from fractional data unlike many other approaches to boosting.

Our staged approach to building mixture models when applied to density estimation is similar to an approach to density estimation suggested by Li and Barron (2000). Li and Barron (2000) provide elegant theoretical results bounding the Kullback-Leibler divergence between an infinite mixture generative density and an approximate finite mixture density. They show that an iterative procedure for the construction of a finite mixture model for density estimation can achieve these bounds. Using the nomenclature of our paper, they show that using a staged mixture modeling approach for density estimation to construct a finite mixture model with parametric components can be guaranteed to approach the generative density. Their results and procedure, however, are limited to the case of density estimation with finite mixtures in which the component models are parametric density models (i.e. having no non-trivial structural component). It would be interesting to extend their theoretical results bounding the Kullback-Leibler divergence to the case of regression and classification models and to the case in which the components have a non-trivial structural component.

There are several other areas for future research. First, it would be useful to demonstrate that the approach yields improvements for other types of component models such as logistic regressions and support vector machines. Second, the SMM approach should be compared to methods for constructing mixture models (other than the backfitting methods and alternative schedules that we used for comparison). Finally, many approaches such as Friedman et al. (1998) regularize their components. It would be useful to experiment with methods for regularizing the mixture weight as well as parameter and structure learning of the components.

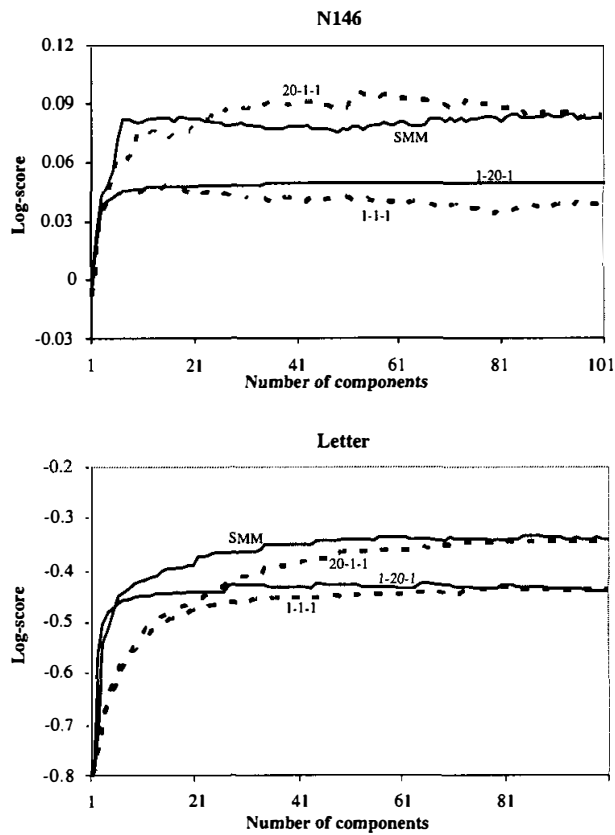

Figure 4: Log-score as a function of the number of components in the learned mixture for N146 (top) and Letter (bottom) using four different learning schedules